\documentclass[11pt,a4paper]{article}
\usepackage{emnlp2021}
\usepackage{times}
\usepackage{latexsym}
\usepackage{amsmath}
\usepackage{url}
\usepackage{booktabs}
\usepackage{multicol}
\usepackage{multirow}
\usepackage{enumitem}
\usepackage{diagbox}
\usepackage{comment}
\usepackage{graphicx}
\usepackage{pifont}
\newcommand{\cmark}{\color{green}{\ding{51}}}%
\newcommand{\xmark}{\color{red}{\ding{55}}}%
\graphicspath{ {./images/} }

\usepackage[T1]{fontenc}

\usepackage[utf8]{inputenc}

\usepackage{microtype}

\title{Model Stability with Continuous Data Updates}

\author{Huiting Liu\thanks{\hspace{1.5mm}Work done while at Apple} \\
  Moloco \\
  \texttt{huiting.liu@moloco.com} \\\And
Avinesh P.V.S\\
  Apple \\
  \texttt{avineshpvs@apple.com} \\\AND
Siddharth Patwardhan \\
  Apple\\
  \texttt{patwardhan.s@apple.com} \\\And
Peter Grasch \\
  Apple\\
  \texttt{pgrasch@apple.com} \\\And
Sachin Agarwal \\
  Apple\\
  \texttt{sachin\_agarwal@apple.com}\\
  }

\date{}

\begin{document}
\maketitle

\begin{abstract}

In this paper, we study the ``stability'' of machine learning (ML) models within the context of larger, complex NLP systems with continuous training data updates.
For this study, we propose a methodology for the assessment of model stability (which we refer to as {\em jitter}) under various experimental conditions. 
We find that model design choices, including network architecture and input representation, have a critical impact on stability through experiments on four text classification tasks and two sequence labeling tasks. 
In classification tasks, non-RNN-based models are observed to be more stable than RNN-based ones, while the encoder-decoder model is less stable in sequence labeling tasks. Moreover, input representations based on pre-trained fastText embeddings 
contribute to more stability than other choices.
We also show that two learning strategies -- {\em ensemble models} and {\em incremental training} -- have a significant influence on stability. We recommend ML model designers to account for trade-offs in accuracy and jitter when making modeling choices.

\end{abstract}

\section{Introduction}
\label{section:introduction}
In industry settings, the data of a machine learning (ML) system are updated frequently to cover the latest trends or new use cases. After a data update, the ML model is retrained to obtain the benefit from data changes. However, the main focus of such processes for a long time has been about the models' accuracy after retraining while ignoring their stability in output distributions.
The table \ref{table:example} illustrates an example of the prediction shift as a measure of instability. 
Given a test set of 100 examples and two different runs of the model M$_{1}$, M$_{2}$ with accuracy 95\% and 96\% for biLSTM and textCNN. 
On the one hand, the textCNN on the right represents the best case, where the model M$_{2}$ corrects 1\% of the model M$_{1}$'s errors (X$_{95}$). 
On the other hand, biLSTM on the left represents the worst case, where M$_{2}$ changes 5\% of the incorrect predictions of model M$_{1}$ to correct (X$_{91}$ \ldots X$_{95}$) and changes 4\% of correct predictions to incorrect (X$_{97}$ \ldots X$_{100}$).
Although the difference in the accuracy of both biLSTM and textCNN is similar, i.e., variance $\pm 1$, the prediction shift can range from 1\% to 9\%. 
A model with such instability can cause a bad user experience in the production environment, where users want their everyday use cases of a product to be stable. 
Even if the behavior for a specific use-case is incorrect, incorrect in a consistent way is preferred over unexpected behavior. 
This type of stability is even more critical in a sensitive field, such as medical ML \citep{Lim3919}, because an unstable ML model may bring severe consequences to the patients and the reputation of medical institutions.   

\begin{table}[t!]
\centering
\small
\begin{tabular}{p{1.5cm}|cc|cc}
\toprule
\multirow{2}{*}{\bf Examples} 
& \multicolumn{2}{c|}{\bf biLSTM} 
& \multicolumn{2}{c}{\bf textCNN} \\
& $M_{1}$  & $M_{2}$ & $M_{1}$ & $M_{2}$ \\
\midrule
$X_{1}$ \ldots $X_{90}$ &  \cmark & \cmark & \cmark & \cmark\\
$X_{91}$  & \xmark & \cmark & \xmark & \xmark\\
$X_{92}$  & \xmark & \cmark & \xmark & \xmark\\
$X_{93}$  & \xmark & \cmark & \xmark& \xmark\\
$X_{94}$  & \xmark & \cmark & \xmark & \xmark\\
$X_{95}$  & \xmark & \cmark & \xmark & \cmark \\
$X_{96}$  & \cmark & \cmark & \cmark & \cmark \\
$X_{97}$  & \cmark & \xmark & \cmark & \cmark \\
$X_{98}$  & \cmark & \xmark & \cmark & \cmark \\
$X_{99}$  & \cmark & \xmark & \cmark & \cmark\\
$X_{100}$  & \cmark & \xmark & \cmark & \cmark\\
\midrule
Accuracy & 95\% & 96\% & 95\% & 96\% \\
\midrule
Variance & 
\multicolumn{2}{c|}{$\pm$1} & 
\multicolumn{2}{c}{$\pm$1}\\
\midrule
Pred. Shift & 
\multicolumn{2}{c|}{9\%} & 
\multicolumn{2}{c}{1\%}\\
\bottomrule
\end{tabular}
\caption{An example of prediction shift of two models in two different runs with the same change in accuracy.}
\label{table:example}
\end{table}

ML models in industry settings are also typically part of a larger system, where the output from one part of the system becomes the input to another.
A consequence of such dependencies among components is that drastic changes in one part of the system directly impact other parts -- potentially altering the system's quality as a whole.
This impact is frequently negative, {\em even if component behavior may have changed positively}.
Such dependencies are considered hidden technical debt in ML systems \cite{sculley2015hidden}. Therefore, a method to measure model stability and guide stability-inducing design choices is desired to address such debt.

Given the above motivations, we want to propose a methodology for the assessment of model stability.
Specifically, the notion of stability we are concerned with here is connected with {\em continuous data updates} (CDUs).
Improvements and maintenance of a complex system involve continuous iterations of its components, which for ML components commonly involves re-training with updated or ``refreshed'' data.
While better training data is expected to improve an individual component's accuracy, we also want a stable distribution from the component output to reduce the possibility of a negative impact on the larger system. 

We assess the stability of multiple model architectures under various conditions and present our findings in the form of best practices for modeling choices in ``stability-sensitive'' settings.
Our focus in this work is on ML methods that are typically used in NLP technologies within larger, complex systems. 
The conclusions of this paper have direct implications on choices made in their design.
Key contributions of this work include (a) a methodology for assessment of model stability with CDUs (and periodic re-training of these models), (b) a study of factors that influence model stability.

Through four different text classification and two sequence labeling tasks, we assess the impact of several key model design decisions on model stability.
The practical consideration is that an ML model designer must account for trade-offs in accuracy and jitter when making modeling choices.
In classification tasks, non-RNN-based structures are observed to be more stable than RNN-based structures, while the encoder-decoder structure is less stable in sequence labeling tasks. Moreover, the pre-trained fastText embeddings \citep{bojanowski2017enriching} have lower jitter than other input representation choices.
Finally, we also learn that using an ensemble of models and incremental training leads to lower jitter, hence greater stability.

\section{Background Research}

For learning algorithms to be stable, it is desirable to have a low variance in the validation error.
However, there are very few quantitative results that analyze the algorithm's stability for changes in the training data.
Many estimates have been proposed in the literature, one of the most prominent ones being cross-validation estimates (e.g., leave-one-out error, k-fold cross-validation). 
The leave-one-out estimate is one such way that measures the variance of the model by running the model \emph{n} times by removing one of the \emph{n} training samples and validating the training example that was deleted.
\citet{rogers1978} first showed that the variance of leave-one-out validation could be upper bounded, which later \citet{DBLP:journals/neco/KearnsR99} called hypothesis stability.
Although hypothesis stability measures change in the learning model with the change in the training set, it does not capture the stability in terms of the change in predictions of the model.
To address this, \citet{DBLP:conf/nips/FardCCG16} first proposed a metric called \emph{Churn}. 
Churn is defined as the expected percentage of prediction difference in the test set between two classifiers. 
For a given model, a fixed gain in accuracy with less churn represents model stability. 
Although recent works \cite{DBLP:conf/nips/FardCCG16, NIPS2016_6316,DBLP:journals/corr/abs-1809-04198} use prediction churn to reduce the potential risk of updating a classifier such that the model remains consistent with the predictions, there is yet no work investigating how small perturbations in training data affect data robustness, which is an internal property of the model structure. 

To improve the stability of models, \citet{DBLP:conf/nips/Vapnik91} uses structural risk minimization for estimating the function based on a complexity penalty. 
This method is similar to regularization, which control the complexity of models by (a) constraining model structure, e.g., limiting the number of hidden layers, (b) influencing the learning, e.g., through control of weight-decay in neural models \cite{NIPS1991_563}, or (c) adjusting pre-processing, e.g., binning and smoothing of the input features. 
Another line of research uses statistical methods like bagging \cite{DBLP:conf/ecai/AndonovaEEP02, DBLP:journals/ml/Breiman96b} to reduce variance without affecting the accuracy of models. 
In effect, this is achieved by taking an average over multiple estimators trained on random samples of the training data. 

Most work on the stability of learning algorithms is in terms of the loss function and translating such properties into uniform generalization.
\citet{DBLP:conf/nips/FardCCG16} build on this notion and address the problem of training consecutive classifiers to reduce the prediction churn by using a Markov Chain Monte Carlo stabilization. 
Later, \citet{DBLP:journals/corr/abs-1809-04198} followed this work using dataset constraints as a part of empirical risk minimization on the classifier's decisions on targeted data sets for low-churn re-training. 
Additionally, \citet{Patrini:2017} proposes using a stochastic matrix capturing the class, flipping them with backward and forward procedural correction \cite{Sukhbaatar:2015}. 
Although various techniques have been used to improve the models' stability, there is no work investigating the model stability with the change in training data.

\section{Measure of Model Stability}
\label{section:measuring-model-stability}

\begin{table*}[t]
	\centering
	\small
	{\renewcommand{\arraystretch}{1.2}
	\begin{tabular}{l@{\hspace{.20cm}} r @{\hspace{.20cm}} r @{\hspace{.20cm}} c @{\hspace{.20cm}} p{4.3cm} @{\hspace{.20cm}} c @{\hspace{.20cm}} p{2.75cm}}
		\toprule
		{\bf Dataset}  & {\bf \# Train}  & {\bf \# Test}  & {\bf \# Classes} & {\bf Example Input} & {\bf Class Output} & {\bf Seq Output} \\
		\midrule 
		CCD &  59,583 & 6681 &  11 &  I have a private loan with \ldots I called to lower my payment \ldots They REFUSED & Student Loan & -\\
		IJCNLP-CF & 3,037 & 500 &  6 & At present not providing snacks evening time & Complaint  & -\\
		Stack Overflow & 35,676 & 4,000 &  20 & convert nickname to formal name in python \ldots formal counterparts using python & Python & - \\
		Reuters-21578 & 6,999 & 2,742 &  77 & USX Corp said proved reserves of oil and natural gas liquids fell \ldots future market improvement may necessitate their closing. & Crude & - \\
		ATIS & 4,978 & 893 & 21 \& 120 & First class fares from Boston to Denver & Airfare & B-class\_type I-class\_type O O B-fromloc O B-toloc\\
		SNIPS & 13,784 & 700 & 7 \& 72  & Book a reservation for a pub serving burritos & Book Restaurant & O O O O O B-restaurant\_type O B-served\_dish \\
		\bottomrule
	\end{tabular}
    }
	\caption{Tasks and datasets for assessment of model's jitter}
	\label{table:datasets}
\end{table*}

We begin our investigation by first defining a measure of model stability.
This measure ({\em jitter}) is the basis of all our experiments.
We use it to compare the impact of various modeling decisions on the stability of the corresponding models.
Note that the notion of ``stability'' here pertains to the changes in the model's behavior observed with continuous updates in the data for training the model (CDUs).


To measure variations in model behavior corresponding to CDUs, jitter must be a defined as a function of several ``versions'' of a specified model $p_\theta$.
Here $p$ represents a specific model architecture with a specific set of hyper-parameters $\theta$, fixed over multiple training and test regimens.
The training data used to train model $p_\theta$ is the experimental variable that we modify across these train-test regimens.
Given a ``base'' training data $D$ and model $p_\theta$, the model is trained $N$ times, each using a version $D_i$ of the base data set $D$. 
The $N$ models, $p_{\theta 1}$, $p_{\theta 2}$, \ldots, $p_{\theta N}$ trained with these data sets, are applied to a test set $X$ producing predictions $Y_i$ corresponding to each trained model $p_{\theta i}$.

$$Y_i = p_{\theta i}(X)$$

\noindent The notion of the difference between a pair of models, $p_{\theta i}$, and $p_{\theta j}$, as first introduced by \citet{DBLP:conf/nips/FardCCG16} as {\em Churn}, is simply a measure of the proportion of data points in $X$ that the two models' outputs differ on (i.e., differences in $Y_i$ and $Y_j$).
Here, we reuse {\em Churn} to define a notion of {\em ``pairwise jitter''}:

\begin{equation}
  \label{eqn:pairwise-jitter-classification}
  J_{i, j}(p_\theta) = \mathrm{Churn}_{i, j}(p_\theta) = \frac{|p_{\theta i}(x) \neq p_{\theta j}(x)|_{x \in X}}{|X|}
\end{equation}

\noindent where $x$ is a data point in dataset $X$, and $p_{\theta i}(x)$, $p_{\theta j}(x)$ are respectively, the predictions of the two models for $x$.
By extending this to encompass all of the models trained on the derived training sets ($D_1$, $D_2$, \ldots, $D_N$), we take an average over {\em pairwise jitter} (\ref{eqn:pairwise-jitter-classification}) over all pairs of models, and establish a more general definition of {\em jitter}:


\begin{eqnarray}
  J(p_\theta) = \frac{\sum_{\forall i,j \in N} J_{i, j}(p_\theta)}{N \cdot (N - 1) \cdot \frac{1}{2} }, \mathrm{where\:} i < j \label{eqn:jitter}
\end{eqnarray}

\noindent which calculates the average probability of a test example having different predictions from $p_\theta$ when $p_\theta$ is retrained for each of the CDUs. 

While the above expression defines jitter for classification models, it is relatively straightforward to extend this notion to sequence labeling, with an update to equation (\ref{eqn:pairwise-jitter-classification}), like so:

\begin{equation}
  \label{eqn:pairwise-jitter-sequence} 
  J_{i, j}(p_\theta) = \frac{|p_{\theta i}(x_t) \neq p_{\theta j}(x_t)|_{x_t \in x | \forall x \in X}}{\sum_{x \in X}|x|}
\end{equation}

\noindent where $x_t$ is the $t^{th}$ item of sequence $x$ in dataset $X$, and $|x|$ represents the length of sequence $x$.
Equation (\ref{eqn:pairwise-jitter-sequence}) measures the proportion of differences in the decisions of two sequence labeling models on every item of each sequence of the dataset.
We use this in equation (\ref{eqn:jitter}) for the general measure of jitter over several model updates.

Note that jitter for a model is not limited to neural architectures and could include any model that can be trained with labeled training data.
It is also important to note that the commonly discussed measure of \emph{variance} in error rate or accuracy is quite different from {\em jitter}.
Jitter measures the disparities in the models' outputs for each individual test example, while variance measures the disparities in the aggregate accuracy or error rate metric across the models. 
For more details about differences see example in {\em Churn} ({\em ``pairwise jitter''}) from section 1.2 of  \citet{DBLP:conf/nips/FardCCG16} and Appendix A. 
Furthermore, we also explain the bounds of jitter and its relation to accuracy and error rate in Appendix B.


\section{Tasks}
\label{section:strategy-and-tasks}

Our major experiments are on four fairly diverse text classification tasks, namely Consumer Complaints (CCD) \citep{cfpb2018consumer}, Stack Overflow\footnote{\url{https://archive.org/download/stackexchange}}, IJCNLP Customer Feedback \citep{plank:2017:IJCNLP} and Reuters-21578\footnote{\url{https://kdd.ics.uci.edu/databases/reuters21578/reuters21578.html}}.
These cover a variety of domains and writing styles, such as informal end-user questions to more formal writing in news stories, allowing us to draw more broadly applicable conclusions from our experiments.
In addition to the varied domains, the tasks have varied text types (e.g. sentence, paragraph, article), sizes and number of classes.
Table~\ref{table:datasets} summarizes the key characteristics of these datasets (details see Appendix C).
While it would be intractable to conduct large scale experiments across all possible natural language tasks, we expect that these data sets align with a large number of common tasks encountered in practice.


To study model stability beyond classification models, we also conduct experiments on two spoken language understanding (SLU) datasets ATIS \citep{atis-cite} and SNIPS \citep{coucke2018snips}.
They include {\em slot filling} (sequence labeling) and {\em intent detection} (text classification), covering basic sequence-to-sequence architectures in Section \ref{section:impact-of-selected-arch} and we present an example of practical implications in Section \ref{section:practical_implications}.

\section{Demonstration of the Effect of Jitter} 

\begin{table*}[t]
\centering
\small
\begin{tabular}{l*{3}{@{\hspace{.15cm}}|c@{\hspace{.15cm}}c}}
\toprule
\multirow{2}{*}{\bf Categories}
& \multicolumn{2}{c|}{\bf Acc of biLSTM with J=1.76} 
& \multicolumn{2}{c|}{\bf Acc of textCNN with J=1.09}
& \multicolumn{2}{c}{\bf Seq Labeling - transformer} \\
& biLSTM$_{1}$ (\%)  & biLSTM$_{2}$(\%)  & textCNN$_{1}$(\%)  & textCNN$_{2}$(\%)  & Token Acc (\%) & Sent Acc \\
\midrule
AddToPlaylist &  100.00 & 100.00 & 100.00 & 100.00 & 90.92 & 66.00 \\
BookRestaurant &  100.00 & 100.00 & 100.00 & 100.00 & 91.56 & 60.00 \\
GetWeather &  95.00  & 97.00 & 96.00 & 98.00 & 95.85 & 72.00 \\
PlayMusic &  96.00  & 97.00 & 97.00 & 97.00 & 87.57 & 72.00 \\
RateBook &  100.00  & 100.00 & 100.00 & 100.00 & 97.04 & 88.00 \\
SearchCreativeWork &  96.00  & 99.00 & 100.00 & 99.00 & 87.96 & 57.00 \\
SearchScreeningEvent &  96.00  & 91.00 & 93.00 & 95.00 & 94.81 & 81.00 \\
\midrule
Overall Class Acc&  97.57 & \bf{97.71} & 98.00 & \bf{98.43} & 92.32 & 70.85 \\
\midrule 
System-Wide Sent Acc &  \bf{69.14} & 69.11 & 69.32 & \bf{69.68} & - & - \\
\bottomrule
\end{tabular}
\caption{Effect of Jitter (J) from the classification component in a two-step ML system}
\label{table:results_snips}
\end{table*}

In this section, before empirically studying model stability, we demonstrate the effect of jitter on a production system. 
For this, we use the SNIPS, a slot labeling and an intent detect task to simulate a two-step ML system.
Let's assume that an existing system for this task uses a biLSTM model for the intent classification and a transformer model for the sequence labeling. 
Due to a regular refresh on the classification training data, we retrain the classifier. 
In the table \ref{table:results_snips}, biLSTM$_{1}$ is the model before retraining, while biLSTM$_{2}$ is the model after retraining. 
The numbers show that a small improvement is introduced on classification test accuracy with this data update. 
However, if we combine the classification accuracy with the sequence labeling accuracy on sentence level in the table \ref{table:results_snips}, the system-wide accuracy (the percentage of cases that both intent and slot labels are correct) drops. As mentioned in Section \ref{section:introduction}, this is because the downstream of the system can't adopt the changes in the output from the re-modeled component. In other words, the biLSTM classifier is not stable with data updates. However, if we replace biLSTM with textCNN and train with these two data versions, as showed by textCNN$_{1}$ and textCNN$_{2}$, the classification accuracy and the system-wide accuracy improve simultaneously. This aligns to the earlier observation in Section~\ref{section:impact-of-selected-arch}, where textCNN tend to be more robust than biLSTM with CDUs. In fact, we can predict the impact of these two classifer architectures on the larger system before-hand by comparing their jitters. As showed on the header of table \ref{table:results_snips}, textCNN has a lower jitter (1.09) than biLSTM (1.76) on the intent classification task. Therefore, improving textCNN with date updates has lower chance to introduce negative impact to the system.  

\section{Experiments}
\label{subsection:simulating-training-data-updates}

With the above example proving that a model with lower jitter is more robust to the overall system with continuous data updates, we can now pursue our primary objective of examining the impact of common ML modeling decisions on stability by devising a series of experiments observing the correspondence between these decisions and jitter.
Largely, we conduct this empirical study over multiple NLP tasks, and over various crucial modeling considerations in designing an ML component (such as, model structure, input representation, etc.).
We lay out the set of common choices available to us for each modeling consideration, design models expressing those choices, and measure corresponding jitter.

The rest of the section is structured as follows: In section~\ref{section:experiment-settings}, we describe continuous data updates (CDUs) and experimental settings. Section~\ref{section:impact-of-selected-arch}, we investigate the impact of model architecture on jitter. 
In section~\ref{section:impact-of-input-rep} and section~\ref{section:training-strategies}, we assess the jitter trade-off of the input representations and training strategies, respectively.



\subsection{Experimental Setup}
\label{section:experiment-settings}
In real-world scenarios, training dataset $D$ could be updated by an ML practitioner for several reasons.
Some common reasons include additional annotated data for general accuracy improvements, fixes to annotation errors in training data discovered in error analysis, or the addition of targeted training data to support new use cases.
For the experimental setting, we can mimic such CDUs by dropping a small $r$\% of data $N$ times from a labeled dataset $D$ to generate $N$ training sets $D_1$, $D_2$, \ldots, $D_N$ of equal size. 
The strategy for selecting data to be dropped could be random, stratified or other sampling types according to how data is refreshed in a task. 
We can then use these $N$ data sets to represent continuous data updates to a ``base'' dataset of the same size.
For our experiment in the remainder of this paper, we set $N=10$ and $r=1$ with stratified sampling on the train sets (train, test instance counts are shown in Table~\ref{table:datasets}; validation set size is 10\% of the train).
These choices are drawn from intuition through common data updates recently conducted in our prior work.

Furthermore, a random seed is fixed across all experiments so that a model will have the same initialization at the start of training. 
We manually tune the hyper-parameters on validation sets (details in Appendix D) and train the selected models to convergence for each data version and then record jitter for each model across these versions. The models' accuracy are very close in these experiments. The focus of these experiments is on stability. Therefore, we only list jitter and variance of accuracy in the reported tables. However, as indicated in Section~\ref{section:practical_implications}, jitter should be used along with other performance metrics in practice. 

\subsection{Impact of Selected Architecture}
\label{section:impact-of-selected-arch}

\begin{table*}[t!]
\centering
\small
\begin{tabular}{l|cc|cc|cc|cc}
\toprule
\multirow{2}{*}{Methods} 
& \multicolumn{2}{c|}{\bf CCD} 
& \multicolumn{2}{c|}{\bf Stack Overflow} 
& \multicolumn{2}{c|}{\bf IJCNLP-CF} 
& \multicolumn{2}{c}{\bf Reuters-21578} \\
& V  & J & V & J & V & J  & V & J \\
\midrule
biLSTM & 1.21 & 9.26 & 0.76 & 10.90 & 1.62 & 21.44 & 0.71 & 9.02 \\
biLSTMAttn & 0.75 & 9.36 & 1.23 & 12.31 & 1.38 & 21.54 & 0.86 & 10.12 \\
biLSTMCNN & 0.59 & 9.80 & 1.31 & 11.83 & 1.80 & 20.04 &  0.41 & 9.23 \\
textcnn & 0.46 & {\bf 7.33} & 0.52 & {\bf 9.78} & 1.82 & {19.27} & 0.47 & {\bf 5.79} \\
transformer & 0.70 & { 8.64} & 1.17 & {10.64} & 1.99 & {\bf 19.13} & 0.63 & {7.86} \\
\bottomrule
\end{tabular}
\caption{Impact of architecture choices on Varianace (V) and Jitter (J) for 10 runs for Classification tasks}
\label{table:results1}
\end{table*}


\begin{table}[t!]
\centering
\small
\begin{tabular}{l|c@{\hspace{.25cm}}c|c@{\hspace{.25cm}}c}
\toprule
\multirow{2}{*}{Methods} 
& \multicolumn{2}{c|}{\bf ATIS} 
& \multicolumn{2}{c}{\bf SNIPS} \\
& V  & J & V & J \\
\midrule
biLSTM  & 0.10 & \textbf{0.98} & 0.30 & \textbf{3.08} \\
biGRU & 0.15 & 1.54 & 0.32 & 4.33\\
biLSTM-CRF & 0.13 & 1.56 & 0.35 & 4.07 \\
transformer & 0.68 & 3.73 & 0.52 & 6.75 \\
biGRU-EncodeDecode-Attn & 0.75 & 9.61 & 0.63 & 10.73 \\
biLSTM-EncodeDecode-Attn & 0.72 & 9.84 & 0.65 & 11.52 \\
\bottomrule
\end{tabular}
\caption{Variance (V) and Jitter (J) of model architectures for sequence labeling tasks}
\label{table:results-seq2seq}
\end{table}


The primary underlying architecture of the model is typically the first and most fundamental choice presented to the ML modeler.
Many choices are available for a task such as text classification, starting from non-neural models (such as SVMs, random forests) to various options in neural models. 
This work investigates standard neural architectures: {\it convolutional neural network, recurrent neural network, attention, and transformer}, since these are more commonly used as building blocks to construct large and complex ML models in recent years.
Specifically, the models we experimented with these components include: 
Text Convolutional Neural Networks (textCNN) \citep{kim2014convolutional}, and 
Bidirectional Long Short Term Memory Networks (biLSTM) \citep{graves2005bidirectional}, biLSTM with Self-Attention (biLSTMAtt) \citep{xie2019self}, biLSTM with Convolutional Neural Networks (biLSTMCNN), and Transformer \citep{vaswani2017attention}. 

In order to minimize the impact of other confounding factors, we fixed most of the shared hyperparameters across all of these architectures with some exceptions, such as learning rate, batch size (details in Appendix D).
In particular, the CNN component configurations are the same in textCNN and biLSTMCNN, as are the biLSTM in the three biLSTM based models. 

The results for this experiment are presented in Table~\ref{table:results1}.
Each column represents one of our four classification tasks, and each cell in the table contains the jitter corresponding to an architecture choice.
At the outset, we observe that these architectures' choices have a very different impact on jitter.
Our key observation is that textCNN and transformer (encoder only) are more stable than those biLSTM based models. 
Especially, textCNN has the lowest jitter on 3 out of 4 tasks and is on par on the 4th. 
This observation suggests that the non-recurrent neural network structures tend to be more stable with CDUs due to its local features as compared to long-range dependencies. 
Furthermore, to understand the impact of the seq2seq architectures, we experiment on two sequence labeling tasks and the results are reported in Table \ref{table:results-seq2seq}. 
We observe that transformer, biLSTM and GRU with encoder-decoder attention frameworks have significantly higher jitter than vanilla RNNs like biLSTM, biGRU in sequence labeling task. 
We believe the encoder-decoder structure to be a major cause of their instability. 

\begin{table}[t]
\centering
\small
\begin{tabular}{l|@{\hspace{.20cm}}c|@{\hspace{.20cm}}c|@{\hspace{.20cm}}c|@{\hspace{.20cm}}c|@{\hspace{.20cm}}c} 
\toprule \bf
\diagbox{Pair}{Overlaps} & 2C  & 3C & 4C &  \textgreater 4C & Total \\
\midrule
biLSTM(-, Att) & 791 & 176 & 26 & 9 & 1002 \\
biLSTM(-, CNN) & 791 & 180 & 33 & 8 & 1012 \\
biLSTM(Att, CNN) & 856 & 183 & 38 & 10 & 1087 \\
(textCNN, biLSTM) & 581 & 91 & 19 & 2 & 693 \\
(textCNN, biLSTMAtt) & 604 & 104 & 30 & 2 & 740 \\
(text, biLSTM)CNN & 652 & 103 & 26 & 2 & 783 \\
(TR, biLSTM) & 675 & 157 & 23 & 3 & 858 \\
(TR, biLSTMAtt) & 682 & 145 & 32 & 10 & 869 \\
(TR, biLSTMCNN) & 690 & 137 & 37 & 7 & 871 \\
(TR, textCNN) & 516 & 80 & 10 & 4 & 610 \\
\bottomrule
\end{tabular}
\caption{Overlapping examples shared across model pairs, on CCD data. TR: transformer (encoder-only)}
\label{table:results6}
\end{table}

To get a better insight into the relationship between jitter and the network components of an architecture, we perform further analysis to understand the experiment results.
We select test examples in the CCD task that show instability with different predictions across ten instances of each architecture and count the overlaps of these test examples across each pair of the models in the table~\ref{table:results6}. 
Firstly, we observed that the two pairs (biLSTM, biLSTMAtt) and (biLSTM, biLSTMCNN) have a very similar amount of overlapping cases across different classes where biLSTM component is the common architecture among them. 
Secondly, among the three pairs containing textCNN, the (textCNN, biLSTMCNN) pair having a CNN component has the largest number of overlaps, especially in cases with instability between two classes. 
Thirdly, the three biLSTM based architectures share close overlaps with the transformer comparing to the (TR, textCNN) pair. These three observations suggest that the number of cases with unstable predictions is closely related to the network components, strengthening our argument that the model architecture plays an essential role in deciding the stability of predictions with CDUs.


\begin{table*}[t]
\centering
\small
\begin{tabular}{l|cc|cc|cc|cc}
\toprule
\multirow{2}{*}{} 
& \multicolumn{2}{c|}{\bf CCD} 
& \multicolumn{2}{c|}{\bf Stack Overflow} 
& \multicolumn{2}{c|}{\bf IJCNLP-CF} 
& \multicolumn{2}{c}{\bf Reuters-21578} \\
& V  & J & V & J \\
\midrule
Basic Models Average (B) & 0.95 & 8.94 $\pm$ 0.96 & 0.76 & 11.09 $\pm$ 1.00 & 0.35 & 20.28 $\pm$ 1.15 & 0.54 & 8.40 $\pm$ 1.66 \\
\midrule
\multicolumn{9}{l}{\textit{Input Representations}} \\[.2ex]
(B) with fastText Embeddings & 0.58 & {\bf 7.36 $\pm$ 0.65} & 0.40 & {\bf 9.84 $\pm$ 0.88} & 0.80 & {\bf 12.93 $\pm$ 0.71} & 0.41 & {\bf 6.53 $\pm$ 1.13} \\
(B) with GloVe Embeddings & 0.83 & 8.29 $\pm$ 0.93 & 1.79 & 11.13 $\pm$ 1.27 & 0.88 & 15.80 $\pm$ 1.13 & 0.57 & 8.47 $\pm$ 1.44 \\
(B) with BERT Embeddings & 0.36 & 9.15 $\pm$ 0.83 & 0.96 & 13.50 $\pm$ 1.79 & 1.69 & 19.86 $\pm$ 3.38 & 0.66 & 7.24 $\pm$ 0.34 \\
\midrule
\multicolumn{9}{l}{\textit{Incremental Training and Ensemble }} \\[.2ex]
(B) with Ensemble & 0.70 & {\bf 4.91 $\pm$ 0.65} & 0.75 & {\bf 6.62 $\pm$ 0.78} & 0.55 & {\bf 12.38 $\pm$ 0.54} & 0.83 & {\bf 4.53 $\pm$ 0.78} \\
(B) with Incremental Training & 0.64 & 6.46 $\pm$ 1.86 & 0.27 & 8.02 $\pm$ 2.21 & 0.60 & 12.92 $\pm$ 4.74 & 0.53 & 5.70 $\pm$ 2.12 \\
\bottomrule
\end{tabular}
\caption{Jitter (J) and Variance (V) for 10 runs averaged across the five selected basic model architectures (B) with different input representations, ensemble, and incremental training}
\label{table:results2}
\end{table*}

\subsection{Impact of Input Representation}
\label{section:impact-of-input-rep}

Another key consideration in ML model development is the choice of input representation. While a neural network model can learn the word embeddings matrix by itself, pre-trained word embeddings are commonly used as input representations to take advantage of the learned semantic representation of words from a large corpus. 
Here, we investigate the effect of such input representations on model jitter.
In our experiments, we use basic models with the self-learned embeddings as baseline. We compare the baseline with two types of commonly used static pre-trained embeddings: GloVe \citep{pennington2014glove} (50 dimensions, 400k vocabulary) and fastText \citep{bojanowski2017enriching} (300 dimensions, 1M vocabulary), and one type of recent high-profile contextual embeddings: BERT \citep{devlin2018bert} (768 dimensions, 30k word-pieces).
One key point to note is that although these input representations' dimensions and vocabulary differ, it is a common practice to compare them as such \citep{joshi-etal-2019-comparison}.
This does not take away from our conclusions.

From our results in Table~\ref{table:results2}, considering jitter, we observe: (a) introducing pre-trained word embeddings leads to lower jitter (i.e., higher stability) in almost all the experiments; (b) fastText embeddings are the best input representation in terms of stability among our four choices; (c) a surprising observation is that BERT, in most cases, induces less stable (high jitter) models.


\subsection{Incremental Training and Ensemble}
\label{section:training-strategies}
In the setting of continuous data updates, typically, prior models are discarded and replaced with new models trained from refreshed or updated data.
Rather than discard or overwrite prior models, in this section we consider two techniques that build upon multiple models.

First we investigate incremental training, which has previously shown stable models while achieving low error rates \cite{Zang14}.
In incremental training, a model is first trained with an initial version of the dataset $D$, and is then retrained (or fine-tuned) with an updated dataset $D_{i}$.
Furthermore, we also investigate ensemble models, which have consistently shown to induce stability to the learning models \cite{Thomas2000}. 
In our ensemble experiments, an ensemble is constructed from five models trained over each of the updated datasets $D_{i}$.
This scenario is akin to keeping the past $N$ versions of the training dataset and training a model to be included in the ensemble.
Output label with most votes is chosen as prediction.

Table~\ref{table:results2} presents the average jitter across the five model architectures.
We can conclude that both ensemble (E) and incremental training (IT) result in lower jitter compared to the baseline model (B).
In all the experiments, ensemble models outperform incremental training.

\section{Model Complexity and Jitter}
\label{section:analysis}

Observing a simple classifier, such as a majority class classifier having 0\% jitter, one may be tempted to believe that simpler models must be more stable.
We try to assess this question through inspecting our experimental results.
To analyze the impact of model complexity on jitter, we use the number of trainable parameters of the model as a proxy for its complexity.
We defer more intricate explorations of complexity to future work.

We take a look at all architectures reported in Section~\ref{section:impact-of-selected-arch} and input representations considered in Section~\ref{section:impact-of-input-rep}.
For this analysis, we consider the models trained on the CCD task.
Recall that, in our experimental setup, GloVe and the self-learned embedding layer both have 50 dimensions, while fastText embeddings have 300 dimensions, and BERT embeddings have 768 dimensions.
For the basic models with a self-learned embedding layer, the embedding layer itself is a matrix of trainable parameters with size {\em (vocabulary x dimensions)}. Table~\ref{table:trainable_p} shows the number of trainable parameters (excluding and including trainable parameters of input representation) along with the corresponding jitter for all models on the CCD task.

\begin{table}[t!]
\centering
\small
\begin{tabular}{lcccc}
\toprule
\multirow{2}{*}{\bf Methods} 
& \multirow{2}{*}{\bf Emb}
& \multicolumn{2}{c}{\bf \#Trainable PRM}  
& \multicolumn{1}{c}{\bf CCD} \\
& & {\bf EX} & {\bf IN} & {\bf J}\\
\midrule
\multirow{4}{*}{biLSTM}  & - & 20k & 2522k & 9.26  \\ 
    & FT & 116k & 116k & 8.11 \\
	 & GL & 20k & 20k & 9.00 \\
	 & BE & 296k & 296k & 9.88 \\
\hline
\multirow{4}{*}{biLSTMAttn} & - & 217k & 2719k & 9.36 \\
 & FT & 473k & 473k & 7.38 \\
	 & GL &  217k & 217k & 8.47 \\
	 & BE & 952k & 952k & 9.63 \\
\hline
\multirow{4}{*}{biLSTMCNN} & - & 328k & 2830k & 9.80 \\
& FT & 584k & 584k & 7.62 \\
	 & GL & 328k & 328k & 8.97 \\
	 & BE & 1063k & 1063k & 9.33 \\
\hline
\multirow{4}{*}{textCNN} & -  & 332k & 2834k & 7.33 \\
 & FT & 588k & 588k & 6.33 \\
	 & GL & 332k & 332k & 6.73 \\
	 & BE & 1067k & 1067k & 7.75 \\
\hline
\multirow{4}{*}{transformer} & -  & 124k & 2626k & 8.64 \\
 & FT & 736k & 736k & 7.34 \\
	 & GL & 124k & 124k & 6.01 \\
	 & BE & 1882k & 1882k & 8.32 \\
\bottomrule
\end{tabular}
\caption{Jitter (J) vs. model complexity -- number of trainable parameters (PRM), excluding (EX), including (IN) input representation in PRM; self-learned embeddings (-), fastText (FT), GloVe (GL) and BERT (BE)}
\label{table:trainable_p}
\end{table}


Our results show no discernible correlation between the number of trainable parameters and jitter, verifying 
\citet{zhang2016understanding}'s suggestion that parameter counting is not a great measure of the model complexity. 
In future work, we plan to investigate other reflections of model complexity, such as degrees of freedom \cite{gao2016degrees} and intrinsic dimension \cite{li2018measuring}, and assess their impact on jitter.

\section{Practical Implications}
\label{section:practical_implications}

\begin{figure}[t]
	\includegraphics[width=0.5\textwidth]{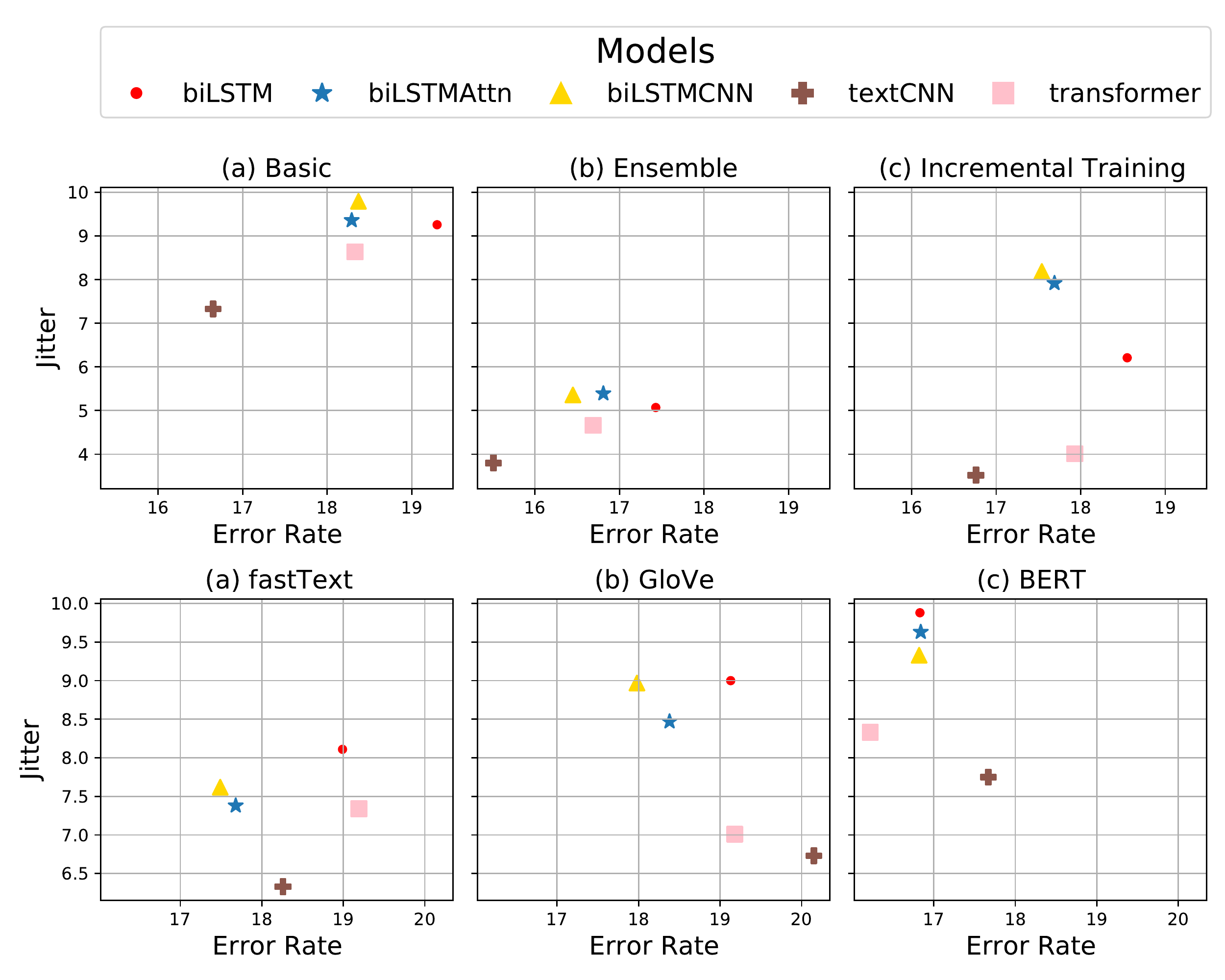}
	\caption{Experiments on  ensemble, incremental training and input representations}
	\label{fig:result1}
\end{figure}

In practice, the ML modeling choices should consider the trade-off between error rate (or accuracy) and jitter.
The plots in Figure~\ref{fig:result1} present the results of our modeling choices on the CCD task.
We plot jitter on the y-axis against error rate on the x-axis.
An ideal model choice would be one that falls in the lower-left corner of the plot, capturing both a low error rate and low jitter.
 
When choosing the input representation for models on this task, fastText may be a better choice even if it does not have the best error rate.
While BERT has lower error rates, models with fastText are observed towards the lower-left corner of our plots. 
Similarly, both ensemble and incremental training have lower jitter and lower error rates consistently.
We recommend to use such plots to identify trade-offs in design choices.

\section{Conclusion}

We investigate a common issue in large complex systems -- that of ML model ``stability'' in the context of continuous data updates.
Specifically, we look into the impact of modeling choices on model stability measured using jitter.
We find that architecture and input representation have a critical impact on jitter.
Non-RNN-based models tend to be more stable in classification tasks than RNN-based ones.
At the same time, the encoder-decoder structure is less stable in sequence labeling tasks.  
Similarly, we observe that pre-trained fastText embeddings are more stable than other input representations.
We also learn that model ensembles and incremental training have lower jitter, hence greater stability.
Of the two, ensemble methods clearly are more stable than incremental training.
Lastly, the practical consideration is that ML model designers must account for trade-offs in accuracy and jitter when designing their model.

We also verify that there is no apparent correlation between jitter and model complexity considering the number of trainable parameters.
The stability of neural models with CDUs is important to understand within complex systems, and this work provides us with tools to understand it better.


\bibliography{jitter_paper}

\begin{thebibliography}{30}
\expandafter\ifx\csname natexlab\endcsname\relax\def\natexlab#1{#1}\fi

\bibitem[{Andonova et~al.(2002)Andonova, Elisseeff, Evgeniou, and
  Pontil}]{DBLP:conf/ecai/AndonovaEEP02}
Savina Andonova, Andr\'{e} Elisseeff, Theodoros Evgeniou, and Massimiliano
  Pontil. 2002.
\newblock A simple algorithm for learning stable machines.
\newblock In \emph{Proceedings of the 15th Eureopean Conference on Artificial
  Intelligence, ECAI'2002, Lyon, France, July 2002}, pages 513--517.

\bibitem[{Bojanowski et~al.(2017)Bojanowski, Grave, Joulin, and
  Mikolov}]{bojanowski2017enriching}
Piotr Bojanowski, Edouard Grave, Armand Joulin, and Tomas Mikolov. 2017.
\newblock Enriching word vectors with subword information.
\newblock \emph{Transactions of the Association for Computational Linguistics},
  5:135--146.

\bibitem[{Breiman(1996)}]{DBLP:journals/ml/Breiman96b}
Leo Breiman. 1996.
\newblock \href {https://doi.org/10.1007/BF00058655} {Bagging predictors}.
\newblock \emph{Machine Learning}, 24(2):123--140.

\bibitem[{CFPB(2018)}]{cfpb2018consumer}
CFPB CFPB. 2018.
\newblock Consumer complaint database check.

\bibitem[{Cotter et~al.(2018)Cotter, Jiang, Wang, Narayan, Gupta, You, and
  Sridharan}]{DBLP:journals/corr/abs-1809-04198}
Andrew Cotter, Heinrich Jiang, Serena Wang, Taman Narayan, Maya~R. Gupta,
  Seungil You, and Karthik Sridharan. 2018.
\newblock \href {http://arxiv.org/abs/1809.04198} {Optimization with
  non-differentiable constraints with applications to fairness, recall, churn,
  and other goals}.
\newblock \emph{CoRR}, abs/1809.04198.

\bibitem[{Coucke et~al.(2018)Coucke, Saade, Ball, Bluche, Caulier, Leroy,
  Doumouro, Gisselbrecht, Caltagirone, Lavril, Primet, and
  Dureau}]{coucke2018snips}
Alice Coucke, Alaa Saade, Adrien Ball, Théodore Bluche, Alexandre Caulier,
  David Leroy, Clément Doumouro, Thibault Gisselbrecht, Francesco Caltagirone,
  Thibaut Lavril, Maël Primet, and Joseph Dureau. 2018.
\newblock \href {http://arxiv.org/abs/1805.10190} {Snips voice platform: an
  embedded spoken language understanding system for private-by-design voice
  interfaces}.

\bibitem[{Devlin et~al.(2018)Devlin, Chang, Lee, and
  Toutanova}]{devlin2018bert}
Jacob Devlin, Ming-Wei Chang, Kenton Lee, and Kristina Toutanova. 2018.
\newblock Bert: Pre-training of deep bidirectional transformers for language
  understanding.
\newblock \emph{arXiv preprint arXiv:1810.04805}.

\bibitem[{Dietterich(2000)}]{Thomas2000}
Thomas~G. Dietterich. 2000.
\newblock Ensemble methods in machine learning.
\newblock In \emph{Proceedings of the First International Workshop on Multiple
  Classifier Systems}, MCS '00, page 1–15, Berlin, Heidelberg.
  Springer-Verlag.

\bibitem[{Fard et~al.(2016)Fard, Cormier, Canini, and
  Gupta}]{DBLP:conf/nips/FardCCG16}
Mahdi~Milani Fard, Quentin Cormier, Kevin~Robert Canini, and Maya~R. Gupta.
  2016.
\newblock \href
  {http://papers.nips.cc/paper/6053-launch-and-iterate-reducing-prediction-churn}
  {Launch and iterate: Reducing prediction churn}.
\newblock In \emph{Advances in Neural Information Processing Systems 29: Annual
  Conference on Neural Information Processing Systems 2016, December 5-10,
  2016, Barcelona, Spain}, pages 3171--3179.

\bibitem[{Gao and Jojic(2016)}]{gao2016degrees}
Tianxiang Gao and Vladimir Jojic. 2016.
\newblock Degrees of freedom in deep neural networks.
\newblock \emph{arXiv preprint arXiv:1603.09260}.

\bibitem[{Goh et~al.(2016)Goh, Cotter, Gupta, and Friedlander}]{NIPS2016_6316}
Gabriel Goh, Andrew Cotter, Maya Gupta, and Michael~P Friedlander. 2016.
\newblock \href
  {http://papers.nips.cc/paper/6316-satisfying-real-world-goals-with-dataset-constraints.pdf}
  {Satisfying real-world goals with dataset constraints}.
\newblock In D.~D. Lee, M.~Sugiyama, U.~V. Luxburg, I.~Guyon, and R.~Garnett,
  editors, \emph{Advances in Neural Information Processing Systems 29}, pages
  2415--2423. Curran Associates, Inc.

\bibitem[{Graves et~al.(2005)Graves, Fern{\'a}ndez, and
  Schmidhuber}]{graves2005bidirectional}
Alex Graves, Santiago Fern{\'a}ndez, and J{\"u}rgen Schmidhuber. 2005.
\newblock Bidirectional lstm networks for improved phoneme classification and
  recognition.
\newblock In \emph{International Conference on Artificial Neural Networks},
  pages 799--804. Springer.

\bibitem[{Hemphill et~al.(1990)Hemphill, Godfrey, and Doddington}]{atis-cite}
Charles~T. Hemphill, John~J. Godfrey, and George~R. Doddington. 1990.
\newblock \href {https://doi.org/10.3115/116580.116613} {The atis spoken
  language systems pilot corpus}.
\newblock In \emph{Proceedings of the Workshop on Speech and Natural Language},
  HLT '90, page 96–101, USA. Association for Computational Linguistics.

\bibitem[{Joshi et~al.(2019)Joshi, Karimi, Sparks, Paris, and
  MacIntyre}]{joshi-etal-2019-comparison}
Aditya Joshi, Sarvnaz Karimi, Ross Sparks, Cecile Paris, and C~Raina MacIntyre.
  2019.
\newblock \href {https://doi.org/10.18653/v1/W19-5015} {A comparison of
  word-based and context-based representations for classification problems in
  health informatics}.
\newblock In \emph{Proceedings of the 18th BioNLP Workshop and Shared Task},
  pages 135--141, Florence, Italy. Association for Computational Linguistics.

\bibitem[{Kearns and Ron(1999)}]{DBLP:journals/neco/KearnsR99}
Michael~J. Kearns and Dana Ron. 1999.
\newblock \href {https://doi.org/10.1162/089976699300016304} {Algorithmic
  stability and sanity-check bounds for leave-one-out cross-validation}.
\newblock \emph{Neural Computation}, 11(6):1427--1453.

\bibitem[{Kim(2014)}]{kim2014convolutional}
Yoon Kim. 2014.
\newblock Convolutional neural networks for sentence classification.
\newblock \emph{arXiv preprint arXiv:1408.5882}.

\bibitem[{Krogh and Hertz(1992)}]{NIPS1991_563}
Anders Krogh and John~A. Hertz. 1992.
\newblock \href
  {http://papers.nips.cc/paper/563-a-simple-weight-decay-can-improve-generalization.pdf}
  {A simple weight decay can improve generalization}.
\newblock In J.~E. Moody, S.~J. Hanson, and R.~P. Lippmann, editors,
  \emph{Advances in Neural Information Processing Systems 4}, pages 950--957.
  Morgan-Kaufmann.

\bibitem[{Li et~al.(2018)Li, Farkhoor, Liu, and Yosinski}]{li2018measuring}
Chunyuan Li, Heerad Farkhoor, Rosanne Liu, and Jason Yosinski. 2018.
\newblock Measuring the intrinsic dimension of objective landscapes.
\newblock \emph{arXiv preprint arXiv:1804.08838}.

\bibitem[{Li et~al.(2020)Li, Sperrin, Ashcroft, and van Staa}]{Lim3919}
Yan Li, Matthew Sperrin, Darren~M Ashcroft, and Tjeerd~Pieter van Staa. 2020.
\newblock \href {https://doi.org/10.1136/bmj.m3919} {Consistency of variety of
  machine learning and statistical models in predicting clinical risks of
  individual patients: longitudinal cohort study using cardiovascular disease
  as exemplar}.
\newblock \emph{BMJ}, 371.

\bibitem[{Patrini et~al.(2017)Patrini, Rozza, Menon, Nock, and
  Qu}]{Patrini:2017}
Giorgio Patrini, Alessandro Rozza, Aditya Menon, Richard Nock, and Lizhen Qu.
  2017.
\newblock \href {https://doi.org/10.1109/CVPR.2017.240} {Making deep neural
  networks robust to label noise: A loss correction approach}.
\newblock In \emph{"Proceedings of Conference on Computer Vision and Pattern
  Recognition"}, pages 2233--2241, "Hawai, USA".

\bibitem[{Pennington et~al.(2014)Pennington, Socher, and
  Manning}]{pennington2014glove}
Jeffrey Pennington, Richard Socher, and Christopher~D Manning. 2014.
\newblock Glove: Global vectors for word representation.
\newblock In \emph{Proceedings of the 2014 conference on empirical methods in
  natural language processing (EMNLP)}, pages 1532--1543.

\bibitem[{Plank(2017)}]{plank:2017:IJCNLP}
Barbara Plank. 2017.
\newblock All-in-1: Short text classification with one model for all languages.
\newblock In \emph{Proceedings of the International Joint Conference on Natural
  Language Processing (Shared Task 4)}, Taipei, Taiwan. Association for
  Computational Linguistics.

\bibitem[{Rogers and Wagner(1978)}]{rogers1978}
W.~H. Rogers and T.~J. Wagner. 1978.
\newblock \href {https://doi.org/10.1214/aos/1176344196} {A finite sample
  distribution-free performance bound for local discrimination rules}.
\newblock \emph{The Annals of Statistics}, 6(3):506--514.

\bibitem[{Sculley et~al.(2015)Sculley, Holt, Golovin, Davydov, Phillips, Ebner,
  Chaudhary, Young, Crespo, and Dennison}]{sculley2015hidden}
David Sculley, Gary Holt, Daniel Golovin, Eugene Davydov, Todd Phillips,
  Dietmar Ebner, Vinay Chaudhary, Michael Young, Jean-Francois Crespo, and Dan
  Dennison. 2015.
\newblock Hidden technical debt in machine learning systems.
\newblock In \emph{Advances in neural information processing systems}, pages
  2503--2511.

\bibitem[{Sukhbaatar et~al.(2015)Sukhbaatar, Bruna, Paluri, Bourdev, and
  Fergus}]{Sukhbaatar:2015}
Sainbayar Sukhbaatar, Joan Bruna, Manohar Paluri, Lubomir Bourdev, and Rob
  Fergus. 2015.
\newblock Training convolutional networks with noisy labels.
\newblock In \emph{"Proceedings of International Conference on Learning
  Representations (ICLR)"}, "San Diego, USA".

\bibitem[{Vapnik(1991)}]{DBLP:conf/nips/Vapnik91}
Vladimir Vapnik. 1991.
\newblock \href
  {http://papers.nips.cc/paper/506-principles-of-risk-minimization-for-learning-theory}
  {Principles of risk minimization for learning theory}.
\newblock In \emph{Advances in Neural Information Processing Systems 4, {[NIPS}
  Conference, Denver, Colorado, USA, December 2-5, 1991]}, pages 831--838.

\bibitem[{Vaswani et~al.(2017)Vaswani, Shazeer, Parmar, Uszkoreit, Jones,
  Gomez, Kaiser, and Polosukhin}]{vaswani2017attention}
Ashish Vaswani, Noam Shazeer, Niki Parmar, Jakob Uszkoreit, Llion Jones,
  Aidan~N Gomez, {\L}ukasz Kaiser, and Illia Polosukhin. 2017.
\newblock Attention is all you need.
\newblock In \emph{Advances in neural information processing systems}, pages
  5998--6008.

\bibitem[{Xie et~al.(2019)Xie, Chen, Gu, Liang, and Xu}]{xie2019self}
Jun Xie, Bo~Chen, Xinglong Gu, Fengmei Liang, and Xinying Xu. 2019.
\newblock Self-attention-based bilstm model for short text fine-grained
  sentiment classification.
\newblock \emph{IEEE Access}, 7:180558--180570.

\bibitem[{Zang et~al.(2014)Zang, Zhang, Zhou, and Guo}]{Zang14}
Wenyu Zang, Peng Zhang, Chuan Zhou, and Li~Guo. 2014.
\newblock \href {https://doi.org/10.1186/2196-1115-1-5} {Comparative study
  between incremental and ensemble learning on data streams: Case study}.
\newblock \emph{Journal Of Big Data}, 1:5.

\bibitem[{Zhang et~al.(2016)Zhang, Bengio, Hardt, Recht, and
  Vinyals}]{zhang2016understanding}
Chiyuan Zhang, Samy Bengio, Moritz Hardt, Benjamin Recht, and Oriol Vinyals.
  2016.
\newblock Understanding deep learning requires rethinking generalization.
\newblock \emph{arXiv preprint arXiv:1611.03530}.

\end{thebibliography}
\bibliographystyle{acl_natbib}

\end{document}


\section*{Appendices}
\label{sec:supplemental}

\subsection{Interpreting Jitter}
\label{subsection:interpreting-jitter}

In this section, we discuss what the jitter computations mean for us and how we might interpret them.
Consider an example hypothetical model $p_\theta$ that has a jitter of 10\% on a classification task.
For updates to 1\% of the training data, this jitter measurement tells us that the output predictions of the model will change for about 10\% of the test data.
This jitter is measured independently of the increase or decrease in accuracy of the model from the training data changes.
A model could potentially see a minimal increase in accuracy but have a {\em high jitter}, indicating that the model is sensitive to training data changes.
In a binary classification setting, the interpretation of this behavior would be that the model tends to flip a large number of output predictions from class $A$ to class $B$, and similarly from class $B$ to class $A$.
Using this measure, and the methodology we describe in the following sections, we can assess various design choices in our ML models that contribute to lower or higher jitter.

\begin{table}[h]
\centering
\small
\begin{tabular}{l|cc}
\hline
Models Pairs  & min (\%) & max (\%)  \\
\hline
\emph{p$_{\theta 1}$} (90\%) - \emph{p$_{\theta 2}$} (91\%) & 1  & 19\\
\emph{p$_{\theta 1}$} (90\%) - \emph{p$_{\theta 3}$} (92\%)  & 2 & 18\\
\emph{p$_{\theta 2}$} (91\%) - \emph{p$_{\theta 3}$} (92\%) & 1  & 17\\
\hline
Avg  (91\% $\pm$ 1) & 1.3 & 18\\
\hline
\end{tabular}
\caption{Minimum (min) and maximum (max) jitter between example model runs}
\label{table:jitter_example}
\end{table}

Before moving on to the next section, we would like to briefly point out the distinction between jitter and accuracy. 
Given three different runs of the model, \emph{p$_{\theta 1}$}, \emph{p$_{\theta 2}$}, \emph{p$_{\theta 3}$}, if the model runs \emph{p$_{\theta 1}$}, \emph{p$_{\theta 2}$} and \emph{p$_{\theta 3}$} are 90\%, 91\%, and 92\% accurate.
In the best case, the model \emph{p$_{\theta 2}$} corrects 1\% of the model \emph{p$_{\theta 1}$}'s errors, \emph{p$_{\theta 3}$} corrects 2\% of the model \emph{p$_{\theta 1}$} and 1\% of model \emph{p$_{\theta 2}$}'s errors. 
In the worst case, \emph{p$_{\theta 2}$} changes 10\% of the incorrect predictions of model \emph{p$_{\theta 1}$} to correct and changes 9\% of correct predictions to incorrect.
Similarly, \emph{p$_{\theta 3}$} changes 8\% of \emph{p$_{\theta 1}$}'s and \emph{p$_{\theta 2}$}'s correct to incorrect, moreover, 10\% of \emph{p$_{\theta 1}$}'s and 9\% of \emph{p$_{\theta 2}$}'s errors to correct, respectively.
Although the accuracy of the models is similar with variance $\pm$1, the models can vary jitter based on best and worst cases.
The table~\ref{table:jitter_example} presents the minimum and maximum change in predictions (jitter) for the example model runs \emph{p$_{\theta 1}$}, \emph{p$_{\theta 2}$}, and \emph{p$_{\theta 3}$}.

\subsection{Bounds for Jitter}
\label{subsection:bounds-for-jitter}

In this subsection, we use the notion of jitter and derive the bounds.

\scalebox{0.85}{\parbox{\linewidth}{%
\begin{equation}
  \label{eqn:pairwise-jitter-classification}
  J_{i, j}(p_\theta) = \mathrm{Churn}_{i, j}(p_\theta) = \frac{|p_{\theta i}(x) \neq p_{\theta j}(x)|_{x \in X}}{|X|}
\end{equation}
}}\\

From the equation (\ref{eqn:pairwise-jitter-classification}), the minimum pairwise jitter across models is the positive change in prediction, which can be defined as:

\scalebox{0.85}{\parbox{.5\linewidth}{%
\begin{eqnarray}
   \label{eqn:jitter-min} 
   \begin{split}
  \min \: & J_{i, j}(p_\theta) = \min (\mathrm{+ve\: change} \oplus  \mathrm{-ve\: change}) \\  
  & \: \quad \quad \quad = |A_{i} - A_{j}| \oplus 0 \\
\min  \: & J(p_\theta) = \frac{\sum_{\forall i,j \in N}|A_{i} - A_{j}|}{N \cdot (N - 1) \cdot \frac{1}{2}}, \mathrm{where\:} i < j
  \end{split}
\end{eqnarray}
}}

\noindent where $A_{i}$ and $A_{j}$ are the corresponding accuracies of the trained model $p_{\theta i}$ and $p_{\theta j}$.
Similarly, the maximum pairwise jitter across models is the maximum change in prediction from one model to another, which can be defined as:

\scalebox{0.85}{\parbox{.5\linewidth}{%
\begin{eqnarray}
   \label{eqn:jitter-max} 
   \begin{split}
  \max \: & J_{i, j}(p_\theta) = \max (\mathrm{+ve\: change} \oplus  \mathrm{-ve\: change}) \\
  & \: \quad \qquad = \min (E_{i}, A_{j}) \oplus  \min (E_{j}, A_{i})\\
  \max \: & J(p_\theta) = \frac{\sum\limits_{\forall i,j \in N}(\min (E_{i}, A_{j}) \oplus  \min (E_{j}, A_{i}))}{N \cdot (N - 1) \cdot \frac{1}{2}}  \\
 & \quad \qquad  \qquad \qquad \qquad \qquad \qquad \mathrm{where\:} i <  j
  \end{split}
\end{eqnarray}
}}

\noindent where $E_i$ and $E_j$ are the respective error rates of the trained model $p_{\theta i}$ and $p_{\theta j}$. 
From equations (\ref{eqn:jitter-min}) and (\ref{eqn:jitter-max}), models with jitter close to the variance improve the comparative model's incorrect predictions. 
However, models whose jitter value is close to the minimum of either average error rate or average accuracy is highly unstable. 

\subsection{Datasets}
The tasks used for the experimentation consists of four classification and two sequence labeling datasets.

\paragraph{Consumer Complaints Database (CCD)}
The Consumer Complaint Database (CCD) \citep{cfpb2018consumer} is a dataset of complaints about consumer financial products provided by the Consumer Financial Protection Bureau.
The CCD  task is to classify paragraphs of text into eleven complaint categories, such as {\em student loan}, {\em credit reporting} that identify the topic of the user complaint. 
This tasks consists of 59,583 training and 6681 test samples.

\paragraph{IJCNLP-CF}
IJCNLP Customer Feedback Analysis dataset \citep{plank:2017:IJCNLP} is a customer feedback sentence classification task into seven categories. In many regards, the data is similar to CCD, but the task is a sentence classification task, with six output classes (covering customer feedback, such as {\em complaint}, {\em request}). For our experiments we use the English dataset of 3,037 training and 500 test samples.

\paragraph{Stack Overflow}
The Stack Overflow Data \footnote{\url{https://archive.org/download/stackexchange}} is a dataset includes an archive of Stack Overflow content, including posts, votes, tags and badges.
The tone and style of the Stack Overflow data is more like that seen on user forums and mailing lists, and the task is a 20-class paragraph classification task about the topic of the post (e.g., {\em Python}, {\em C\#}) and contains 35,676 training and 4,000 test samples. 

\paragraph{Reuters}
Reuters\footnote{\url{https://kdd.ics.uci.edu/databases/reuters21578/reuters21578.html}} is much more formal text from newspaper stories, where the task is to classify these paragraphs into 77 topic classes, such as {\em cocoa}, {\em reserves}, {\em rice}, {\em rubber}, {\em heat}, {\em income} etc. The dataset consists of 6,999 training and 2,742 test samples.

\paragraph{ATIS}
Airline Travel Information Systems (ATIS) \citep{atis-cite} is a widely used spoken language understanding dataset containing audio recordings and corresponding manual transcripts about people tasking for flight information and travel reservations. The data consists of 21 unique intent categories (e.g., {\em flight\_time}, {\em airfare},  etc.) and corresponding slot labels. 
The data is split into 4,978 & 893 train and test sets respectively.

\paragraph{SNIPS}
SNIPS \citep{coucke2018snips} is a crowdsourced natural language corpus used to benchmark the performance of voice assistants. It consists of queries distributed among 7 intent categories, such as {\em SearchCreativeWork}, {\em GetWeather}, {\em BookRestaurant}, {\em PlayMusic} etc and corresponding slot labels. The dataset consists of 13,784 & 700 train and test samples, with 100 queries per intent in test set.

\subsection{Experimental Setup}

\section*{Model Hyperparameters}
After manual tuning, we found optimal hyperparameter values that are task-specific; in this subsection, we provide details about the hyperparameters we used across our experiments. All the experiments were conducted on 1 GPU and 1 CPU. 

\subsection*{Selected Architectures - Classification}

\subsection*{biLSTM}
\begin{itemize}[noitemsep,topsep=0pt]
\item Batch size: 37 (CCD, Stack Overflow), 12 (IJCNLP-CF), 20 (Reuters-21578)
\item Learning rate: 0.5
\item Dropout: 0.5
\item Epochs: 15
\item LSTM size: 128 
\item Optimizer: sgd
\end{itemize}

\subsection*{biLSTMAtt}
\begin{itemize}[noitemsep,topsep=0pt]
\item Batch size: 37 (CCD, Stack Overflow), 12 (IJCNLP-CF), 20 (Reuters-21578)
\item Learning rate: 0.5
\item Dropout: 0.5
\item Epochs: 15
\item LSTM size: 128 
\item Optimizer: sgd
\item Attention Heads: 1
\item Attention Dimensions: 128
\item Feed Forward Network Dimensions: 128
\end{itemize}

\subsection*{biLSTMCNN}
\begin{itemize}[noitemsep,topsep=0pt]
\item Batch size: 37 (CCD, Stack Overflow), 12 (IJCNLP-CF), 20 (Reuters-21578)
\item Learning rate: 0.5
\item Dropout: 0.5
\item Epochs: 15
\item LSTM size: 128 
\item Optimizer: sgd
\item CNN filters: 32
\item CNN filter sizes: 3, 4, 5
\end{itemize}

\subsection*{textCNN}
\begin{itemize}[noitemsep,topsep=0pt]
\item Batch size: 37 (CCD, Stack Overflow), 12 (IJCNLP-CF), 20 (Reuters-21578)
\item Learning rate: 0.001
\item Dropout: 0.5
\item Epochs: 15
\item Optimizer: adam
\item CNN Filters: 32
\item Filter Sizes: 3, 4, 5
\end{itemize}

\subsection*{transformer}
\begin{itemize}[noitemsep,topsep=0pt]
\item Batch size: 37 (CCD, Stack Overflow), 12 (IJCNLP-CF), 20 (Reuters-21578)
\item Learning rate: 1.0
\item Dropout: 0.5
\item Epochs: 15
\item Optimizer: adadelta
\item Attention Heads: 8
\item Attention Dimensions: 512
\item Feed Forward Network Dimensions: 128
\end{itemize}

\subsection*{Selected Architectures - Sequence Labeling}

\subsection*{biLSTM}

\begin{itemize}[noitemsep,topsep=0pt]
\item Batch size: 32
\item Learning rate: 0.001
\item Dropout: 0.5
\item Epochs: 8
\item LSTM Size: 128 
\item Optimizer: rmsprop
\end{itemize}

\subsection*{biGRU}
\begin{itemize}[noitemsep,topsep=0pt]
\item Batch size: 32
\item Learning rate: 0.001
\item Dropout: 0.5
\item Epochs: 8
\item Unit Size: 128 
\item Optimizer: rmsprop
\end{itemize}

\subsection*{biLSTM-CRF}
\begin{itemize}[noitemsep,topsep=0pt]
\item Batch size: 32
\item Learning rate: 0.001
\item Dropout: 0.5
\item Epochs: 8
\item Optimizer: rmsprop
\end{itemize}

\begin{table*}[t]
\centering
\begin{tabular}{p{8.3cm}|m{2cm}|m{2.2cm}|m{0.7cm}|m{0.7cm}|m{0.7cm}}
\toprule
\bf Input Text & \bf True label & \bf Model & \bf M$_{1}$ & \bf M$_{2}$ & \bf M$_{3}$ \\ 
\midrule
\multirow{5}{*}{\parbox[t]{8.3cm}{
\ldots to open up a checkin account with my bank and i couldnt becasue someone messed uo my credit getting things in they name steal my idenity i just turn yy old 2015 theres no way i could of got \ldots i need my credit score cleared}}
& \multirow{5}{*}{CR} & biLSTM & BAC & BAC & CR\\
& & biLSTMAttn & CR & BAC & CC\\
& & biLSTMCNN & CR & CR & BAC \\
& & textCNN & CR & CR & CC \\
& & transformer & CR & CC & BAC \\
\midrule
\multirow{5}{=}{\setlength\parskip{\baselineskip}%
i entered a discount settlement agreement and paid the debt as agreed \ldots would update my credit report and remove the lien from the property i have called xxx \ldots still has not been updated on my credit report}
& \multirow{5}{*}{M} & biLSTM  & M & DC & CR\\
& & biLSTMAttn & M & M & DC \\
& & biLSTMCNN & M & M & DC\\
& & textCNN & M & M & DC \\
& & transformer & DC & DC & M \\
\midrule
\multirow{5}{=}{\setlength\parskip{\baselineskip}%
sought medical service, whose doctor prescribed medical prescription the product was defective, xxx was advised not to submit further debits to my account. suntrust was advised to block such payments, however, suntrust allowed payments}
& \multirow{5}{*}{BAS} & biLSTM  & DC & DC & BAS\\
& & biLSTMAttn & DC & DC & CC\\
& & biLSTMCNN & DC & DC & BAS\\
& & textCNN & BAS & BAS & DC  \\
& & transformer & DC & DC & DC \\
\midrule
\multirow{5}{=}{\setlength\parskip{\baselineskip}%
 i sent money through moneygram \ldots the agent said to me moneygram has a new policy to request photocopy of id  \ldots what can moneygram do to ensure me that i will not be an identity theft victim}
& \multirow{5}{*}{MT} & biLSTM  & MT & M & BAS\\
& & biLSTMAttn & MT & MT & DC\\
& & biLSTMCNN & MT & MT & DC\\
& & textCNN & MT & MT & DC  \\
& & transformer & MT & MT & MT \\
\midrule
\multirow{5}{=}{\setlength\parskip{\baselineskip}%
finance company is unwilling to work with me in reducing payments to something that is actually reasonable}
& \multirow{5}{*}{SL} & biLSTM  & CL & CL & DC\\
& & biLSTMAttn & CL & CL & DC\\
& & biLSTMCNN & CL & PL & DC\\
& & textCNN & DC & DC & CL \\
& & transformer & CL & CL & DC \\
\bottomrule
\end{tabular}
\caption{Examples of change in test predictions across models causing jitter, Labels CR: Credit reporting, DC: Debt collection. BAS: Bank account or service, CC: Credit card, M: Mortgage, MT: Money Transfer, CL: Consumer Loan, SL: Student Loan, PL: Payday Loan}
\label{table:examples}
\end{table*}

\subsection*{transformer}
\begin{itemize}[noitemsep,topsep=0pt]
\item Batch size: 32
\item Dropout: 0.1
\item Epochs: 50
\item Optimizer: Adam optimizer with a custom learning rate scheduler according to \cite{vaswani2017attention}
\item Attention Heads: 8
\item Attention Dimensions: 512
\item Feed Forward Network Dimensions: 128
\end{itemize}

\subsection*{biLSTM-EncoderDecoder-Attn}
\begin{itemize}[noitemsep,topsep=0pt]
\item Batch size: 64
\item Learning rate: 0.001
\item Dropout: 0.5
\item LSTM Size: 128 
\item Epochs: 50
\item Optimizer: Adam
\end{itemize}

\subsection*{biGRU-EncoderDecoder-Attn}
\begin{itemize}[noitemsep,topsep=0pt]
\item Batch size: 64
\item Learning rate: 0.001
\item Dropout: 0.5
\item Unit Size: 128 
\item Epochs: 50
\item Optimizer: Adam
\end{itemize}

We take the prediction results of four basic model architectures on Consumer Complain Database from Table 1. 



\subsection{Example Cases}
Table \ref{table:examples} illustrates some of the change in prediction results of five basic model architectures on Consumer Complain Database (CCD). $M_1$, $M_2$, $M_3$ are three instances of these basic models trained on three versions of the CCD dataset.

\bibliography{jitter_paper}
\bibliographystyle{acl_natbib}